# Translation-Invariant Representation for Cumulative Foot Pressure Images

Shuai ZHENG†, Kaiqi HUANG†, and Tieniu TAN†

†National Laboratory of Pattern Recognition, Chinese Academy of Science
No.95 Zhongguancun East Road, Haidian District, Beijing, China
E-mail : †{szheng, kqhuang, tnt}@nlpr.ia.ac.cn

**Abstract** Human can be distinguished by different limb movements and unique ground reaction force. Cumulative foot pressure image is a 2-D cumulative ground reaction force during one gait cycle. Although it contains pressure spatial distribution information and pressure temporal distribution information, it suffers from several problems including different shoes and noise, when putting it into practice as a new biometric for pedestrian identification. In this paper, we propose a hierarchical translation-invariant representation for cumulative foot pressure images, inspired by the success of Convolutional deep belief network for digital classification. Key contribution in our approach is discriminative hierarchical sparse coding scheme which helps to learn useful discriminative high-level visual features. Based on the feature representation of cumulative foot pressure images, we develop a pedestrian recognition system which is invariant to three different shoes and slight local shape change. Experiments are conducted on a proposed open dataset that contains more than 2800 cumulative foot pressure images from 118 subjects. Evaluations suggest the effectiveness of the proposed method and the potential of cumulative foot pressure images as a biometric.

## 1. Introduction

Walking can be considered as an inherent attribute of a person: each person has his/her own unique limb movement pattern and ground reaction force during walking [11]. This attribute allows one to identify a person. Previous works [19, 20] have developed person identification system using the unique limb movement pattern, which is called gait. However, recent gait recognition systems do not work well since there are many challenges in practice, such as illumination change in environment, clothing variation for the same individual and different camera viewpoint for probe and gallery data. Previous biomedical, robotic and forensic studies [8, 12, 14] have proposed two kinds of pedestrian recognition system based on ground reaction force information. One is utilizing barefoot print image, another is based on 1-D ground reaction force signal. Although the uniqueness of barefoot has been proved [12], but shoes-off recognition process is not suitable for most scenarios [2]. Besides, 1-D ground reaction force signal [14] does not provide enough information to recognize individual, although it seems practical to help robot to distinguish pedestrian and other objects. In this paper, we propose a pedestrian recognition system using cumulative foot pressure images. Cumulative foot pressure image contains cumulative spatial and temporal force information during one gait cycle [19, 18], which may help it handling the difficulties in recognizing different shoes-wearing individual.

For some security scenarios such as jailhouse security system, bathhouse, entrance of public transportation, and entrance of Japanese house, camera-based recognition system does not work well since it suffers from various problems such as privacy, difficult extreme environment and crowd backgrounds. The proposed recognition system using cumulative foot pressure images can be deployed in these scenarios, taking place of camera-based system. Different from gait, cumulative foot pressure image is acquired by floor pressure sensing device [14] rather than camera. Hence there is no light illumination change or different camera viewpoint

problem for

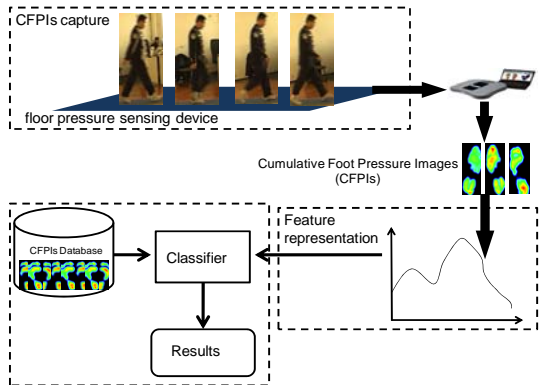

Figure 1. Overview of pedestrian identification system using cumulative foot pressure images

it. Besides, cumulative foot pressure images can help to identify the same individual with different clothes since it doesn't rely on the appearance of individual. Furthermore, it may be more suitable for those specific scenarios. What's more important is that it does not have privacy problem since it does not need to take extra information such as appearance of individual.

In this paper, we propose recognition task based on cumulative foot pressure images: verification (same or not, given two cumulative foot pressure image set). Fig. 1 illustrates the overview process of the proposed pedestrian recognition system using cumulative foot pressure images. When a person walks through foot pressure measurement floor, cumulative foot pressure images are acquired. Based on the acquired cumulative foot pressure image set, there are two major components to achieve the proposed two recognition tasks: cumulative foot pressure image representation and classification. The extracted descriptor is required to be not only discriminative but also invariant to apparent changes due to different shoes and noise. Based on the representation, the classification should be robust to shape variations. As shown in Fig. 1, these requirements render cumulative foot pressure image recognition a challenging problem.

Currently, descriptor-based approaches [3, 4, 6, 7] are proved to be effectiveness in many representative challenge recognition dataset such as MINIST [6], and Caltech101 [6]. Many low-level descriptors such as SIFT or HOG are designed to be invariant to minor translations of input images. However, these low-level descriptors are not practical for the proposed system since they lose a lot of structure information (in terms of object part) which is crucial for shoes-invariant cumulative foot pressure image recognition. Several authors have proposed unsupervised methods to learn image descriptors based on sparse/overcomplete decomposition [10], but they have not been proved to be local invariance. Many successful object recognition proposals have appeared for unsupervised learning of locally-invariant descriptors, which also use sparsity criteria [21]. Besides, the sparse coding and locality-constrained sparse coding have been proved to be efficient for more challenge PASCAL VOC object recognition dataset [22]. Hence sparsity and locality criteria have the potential in our proposed system.

Our goal is to learn a feature representation model for the cumulative foot pressure image which preserves discriminative local and global structure information. Hence, we propose a discriminative hierarchical locality-constrained sparse coding scheme for representing cumulative foot pressure image. The whole representation scheme is consist of two steps, one is dictionary learning, and another is hierarchical coding.

The remainder of this paper is organized as follows. In Section 2, an overview description of feature representation for cumulative foot pressure image is given. In Section 3, sparse coding algorithm and its variation algorithms are described. In Section 4, a hierarchical locality-constrained sparse coding for cumulative foot pressure image is presented. In Section 5, experimental results on the dataset contains barefoot cumulative foot pressure image and shoes-wearing cumulative foot pressure image are reported, and corresponding analysis is presented. Conclusion and future works are given in Section 6.

## 2. Overview of Pedestrian Recognition using Cumulative Foot Pressure Image

As Fig. 2 illustrate, similar to [22], we propose a hierarchical locality-sparse coding for cumulative foot pressure image. The differences between [17] and this paper are hierarchical coding strategies and pooling operations.

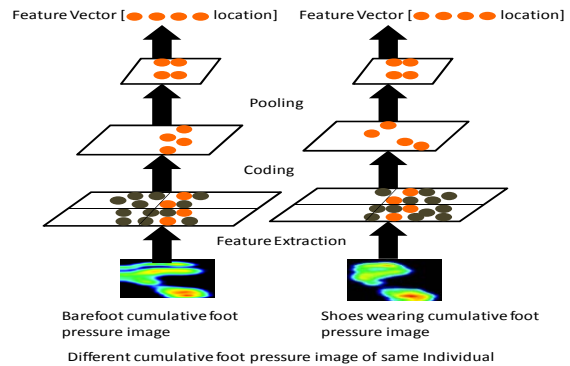

Figure 2. Hierarchical locality-constrain sparse coding scheme.

## 3. Hierarchical Locality-constrain Sparse Coding Scheme

### 3.1 Sparse coding

Sparse coding algorithms model an input cumulative foot pressure image $x \in R^m$ using a linear combination of basis functions that are columns of the dictionary matrix $D \in R^{m \times n}$, using coefficients $z \in R^n$, with $n > m$. Previous sparse coding algorithms have been proposed in the literature and in this work we focus on the following convex formulation:

$$L = \min \frac{1}{2}\|x - Dz\|_2^2 + \lambda \sum_i |z_i| \qquad (1)$$

The basic idea of sparse coding is to minimize the same objective of Eqn. 1 alternatively over coefficients $z$ for a given dictionary $D$, and then over $D$ for a given set of $z$.

### 3.2 Locality-constrained sparse coding

As described in [22], Locality-constrained linear coding (LLC) introduce locality constraint instead of the sparsity constraint only, which leads to several favorable properties. Sepcifically, the LLC coding uses the following criteria:

$$L = \min \frac{1}{2}\|x - Dz\|_2^2 + \lambda \sum_i |d_i \odot z_i| \qquad (2)$$

where $\odot$ denotes element-wise multiplication, di is the locality adaptor that gives different freedom for each basis vector proportional to its similarity to the input cumulative foot pressure image. Different from [22], we set $d_i$ as

$$d_i = dist(x_i, B) \qquad (3)$$

Hence, we can derive analytical solution of (2), and obtain the solution as

$$\begin{cases} \tilde{z}_i = (Z_i + diag(d)) \backslash 1 \\ z_i = \tilde{z}_i / 1^T \tilde{z}_i, \end{cases} \qquad (4)$$

Where $Z_i = (B - 1x_i^T)(B - 1x_i^T)^T$ denotes the data covariance matrix.

## 4. Hierarchical Locality-constrained Sparse Coding

### 4.1 Hierarchical Locality-constrained Sparse Coding using Gaussian weights

There are two terms in (2), one is optimizing the reconstruction error, another one is minimizing the regularization penalty terms. Considering cumulative foot pressure image may present different appearance, only optimizing the two terms is not enough to reflect the structure information. Hence we propose to let z contain the correlation with its K neighbor patches. Supposing z follows Gaussian distribution, we modified (2), and obtain a new optimization formulation:

$$L = \min \frac{1}{2}\|x - Dz\|_2^2 + \lambda \sum_i |w_i z_i| \qquad (5)$$

Where $w_i$ is the Euclidean distance between $z_i$ and its nearest Gaussian distribution center.

### 4.2 Discriminative HLSC

Sparse coding, locality-constrained sparse coding, hierarchical locality-constrained sparse coding are all unsupervised feature selection methods. Previous studies show the effectiveness of combination of unsupervised method with discriminative model [23]. Inspired by them, we proposed fisher linear discriminative analysis model to extract the discriminative components after hierarchical linear-constrained sparse coding.

Considering the representation z, which is the feature vectors obtained from hierarchical linear-constrained sparse coding, we also collect the corresponding label y. fisher linear discriminative analysis give a dimensionality reduced components z'. Based on the process, we can achieve both discriminative and approximated lowest-reconstruction-error feature representation z'.

## 5. Experimental Results

In this section, we report results based on two proposed cumulative foot pressure dataset: barefoot dataset and shoes-invariant dataset. As far as we know, there are no previous public accessible dataset for cumulative foot pressure images. One contains all barefoot cumulative foot pressure images and its corresponding gender and identity label. It is collected from 88 subjects, including 68 male and 20 female. Another sub dataset contains barefoot cumulative foot pressure images, wearing cloth shoes one, wearing sports shoes one and wearing leather shoes one. It is collected from 30 male subjects. These subjects came from different areas and they are constituted with different age groups from 22 to 50. Hence the experimental results are not limited to a specific group of people.

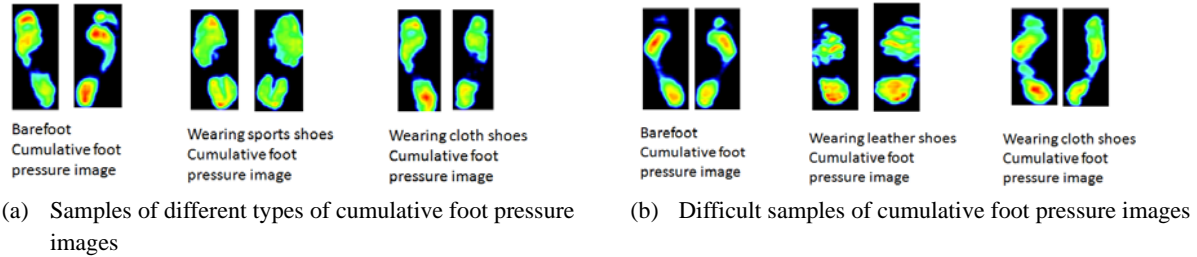

(a) Samples of different types of cumulative foot pressure images

(b) Difficult samples of cumulative foot pressure images

Figure. 3 Samples of cumulative foot pressure dataset

Pedestrian identification system often uses false acceptance rate (FAR) and false rejection rate (FRR) as performance evaluation criteria.

Considering the classifier factor, we only compare the results of different feature representation methods, hence we choose to use multi-class support vector machine (SVM) as our classifier. And the regularization coefficient is set as 0.1 for different feature representation methods. In sparse coding framework, we used only Dense SIFT [4] descriptor as our low-level descriptor, throughout the experiment. In out setup, the SIFT descriptor are extracted from patches densely located by every 4 pixels on the image, under the scale, 4x4 respectively. The dimension of each SIFT descriptor is 128. During feature representation learning process,

PCA is our benchmark considering the similar method described in [1,11]. We set reduced dimensionality as 200. Restricted Boltzmann Machine [5] is a popular deep learning feature selection method, which is proved to be effectiveness in digital recognition. Still, we set reduced dimensionality as 200. We choose to compare our works with it since it is a related representative of unsupervised learning method. DHSC is our proposed method. Where dictionary size is set as 128 and regularization penalty term is set as 0.15.

Table. 1, 2, 3 show the experimental results of comparison between different methods. Generally speaking, from the results, we can see the effectiveness of sparsity in representing the structure information. Discriminative hierarchical linear-constraint sparse coding method outperforms the PCA+LDA and RBM+LDA methods, since the sparsity penalty terms in (5) ensure the sparsity of representation while preserve the important structure information from cumulative foot pressure images.

However, the shoes-invariant cumulative foot pressure image recognition is still a difficult problem. All three representative methods do not well handle the problem that recognizing shoes-wearing subject when registered one is barefoot one. As fig. 3 illustrate, the appearance of barefoot cumulative foot pressure images are different from that of shoes wearing one. Even for human, we may not distinguish these types of data easily. All three methods and other possible traditional computer vision methods are all focus on modeling the appearance information of image. Future works need to take into account of other factor and other correlated modal like gait pose images together.

Fig. 4 illustrate the important components of a given cumulative foot pressure image. These components are the key of recognizing the corresponding pedestrian and distinguish it with the cumulative foot pressure images of others.

Table 1. Experiments are conducted under the condition that bare feet are used in registering and testing.

|     | PCA [1,11]+LDA | RBM [5]+LDA | DHSC |
|-----|----------------|-------------|------|
| FRR | 0.2%           | 0.2%        | **0.1%** |
| FAR | 0.6%           | 1.0%        | **0.9%** |

Table 2. Experiments are conducted under the condition that bare feet are used in registering while shoes wearing ones are used in testing.

|     | PCA[1,11]+LDA | RBM[5]+LDA | DHSC |
|-----|---------------|------------|------|
| FRR | 28.6%         | 22.0%      | **19.0%** |
| FAR | 1.3%          | 1.0%       | 1.0% |

Table 3. Experiments are conducted under the condition that wearing shoes feet are used in registering while shoes wearing ones are used in testing.

|     | PCA [1,11]+LDA | RBM [5]+LDA | DHSC |
|-----|----------------|-------------|------|
| FRR | 0.5%           | 0.4%        | **0.2%** |
| FAR | 2.0%           | 1.0%        | **0.02%** |

## 6. Conclusion and Future Works

In this paper, we discuss the potential of cumulative foot pressure image as a biometric and present a possible solution to its challenge in shoes-invariant recognition. However, the results prove the effectiveness of the proposed method. Discriminative hierarchical linear-constraint sparse coding method achieves best performance comparing with other potential methods due to that it model both the structure and appearance information of a given cumulative foot pressure image. Besides, it sounds reasonable that the key components are more discriminative. Because these components of a given cumulative foot pressure image are corresponding to one gait sequence [18].

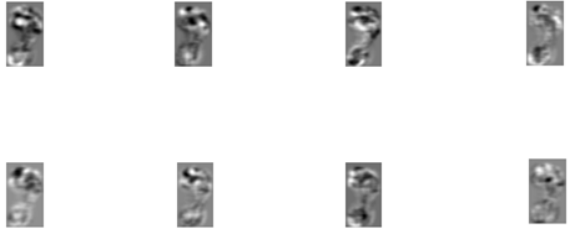

Figure. 4 Important components of cumulative foot pressure in discriminative hierarchical linear-constrained sparse coding

However, our work is only an initial work on cumulative foot pressure images. Future works are needed in at least four directions. One is that the shoes-invariant pedestrian recognition using cumulative foot pressure images. The performance of our proposed method does not meet the practical needs. Besides, the relations between cumulative foot pressure images and footprint are needed to be mentioned. Once the relations are found, some forensic applications such as suspect identification would be more reliable and more efficient. Thirdly, the relations between cumulative foot pressure images and gait are needed to be studied. Based on the studies, we can develop a more practical fusion-based pedestrian identification system. Finally but not least, a large and standard open dataset with specific evaluation criteria are needed. In this paper, we have contributed to the first one and fourth one.